\documentclass{article}
\usepackage{ijcai19}

\usepackage{times}
\usepackage{xcolor}

\usepackage{array}
\usepackage{amsmath}
\usepackage{amssymb}
\usepackage{color}
\usepackage{multicol}
\usepackage{multirow}
\usepackage{array}
\usepackage{booktabs}
\usepackage{balance}
\usepackage{arydshln}
\usepackage{url}

\usepackage{xspace}

\newcommand{\etal}{\emph{et al.,}\xspace}
\newcommand{\eg}{\emph{e.g.,}\xspace}
\newcommand{\ie}{\emph{i.e.,}\xspace}

\DeclareMathOperator*{\argmax}{arg\,max}

\usepackage{helvet}
\usepackage{courier}

\usepackage{bm}
\usepackage{color}

\usepackage{arydshln}

\usepackage{amsfonts}
\usepackage{graphicx} 
\usepackage{graphics}
\usepackage{standalone}

\usepackage{caption}
\usepackage{subfigure}
\usepackage{balance}

\DeclareSymbolFont{extraup}{U}{zavm}{m}{n}
\DeclareMathSymbol{\varheart}{\mathalpha}{extraup}{86}
\DeclareMathSymbol{\vardiamond}{\mathalpha}{extraup}{87}

\title{Learning to Refine Source Representations for Neural Machine Translation}

\author{
Xinwei Geng$^\varheart$\thanks{Work done when interning at Tencent AI Lab.} ~~~ Longyue Wang$^\spadesuit$ ~~~ Xing Wang$^\spadesuit$ ~~~ Bing Qin$^\varheart$ ~~~ Ting Liu$^\varheart$ ~~~ Zhaopeng Tu$^\spadesuit$
\affiliations
$^\varheart$Harbin Institute of Technology   ~~~~~~~~~~~~~~~~~~~~~~~~~~~~~~~ $^\spadesuit$Tencent AI Lab ~~~~~~~~~~~~~~~~~~~~~~~
\emails
\normalsize $^\varheart$\{xwgeng, qinb, tliu\}@ir.hit.edu.cn  ~~~~~~~~~~~~~~~~~~ $^\spadesuit$\{vinnylywang, brightxwang, zptu\}@tencent.com
}

\begin{document}
\maketitle

\begin{abstract}

Neural machine translation (NMT) models generally adopt an encoder-decoder architecture for modeling the entire translation process. 
The encoder summarizes the representation of input sentence from scratch, which is potentially a problem if the sentence is ambiguous.
When translating a text, humans often create an initial understanding of the source sentence and then incrementally refine it along the translation on the target side. Starting from this intuition, we propose a novel {\em encoder-refiner-decoder} framework, which dynamically refines the source representations based on the generated target-side information at each decoding step. Since the refining operations are time-consuming, we propose a strategy, leveraging the power of reinforcement learning models, to decide when to refine at specific decoding steps. Experimental results on both Chinese--English and English--German translation tasks show that the proposed approach significantly and consistently improves translation performance over the standard encoder-decoder framework. Furthermore, when refining strategy is applied, results still show reasonable improvement over the baseline without much decrease in decoding speed.
\end{abstract}

\section{Introduction}
The encoder-decoder framework has achieved promising progress in sequence generation tasks including dialog system~\cite{vinyals2015neural,li2017adversarial}, question answering~\cite{xiong2016dynamic,chen2017reading} as well as machine translation~\cite{sutskever2014sequence,Bahdanau:2015:ICLR}. In neural machine translation (NMT), the encoder summarizes the source sentence into a vector representation, and the decoder generates the target sentence word-by-word from the vector representation. Besides, the attention mechanism~\cite{Bahdanau:2015:ICLR} dynamically select parts of the source representations according to its relevance to the next target word.

However, the encoder summarizes the representation of input sentence from scratch, which is potentially a problem if the sentence is ambiguous.
Like a human translator, the encoding process is analogous to reading a sentence in the source language and summarizing its meanings (\ie source representations) for generating the equivalents in the target language. 
When humans translate some complex sentences, they generally create an initial understanding on the source sentence (which may be wrong), and then incrementally {\em refine} the understanding based on the partial translation they have generated~\cite{hayes1986writing}.
As seen, it is difficult even for humans to translate text based on unchanged understanding in a single pass. 

\begin{figure}[t]
\centering
\includegraphics[width=0.38\textwidth]{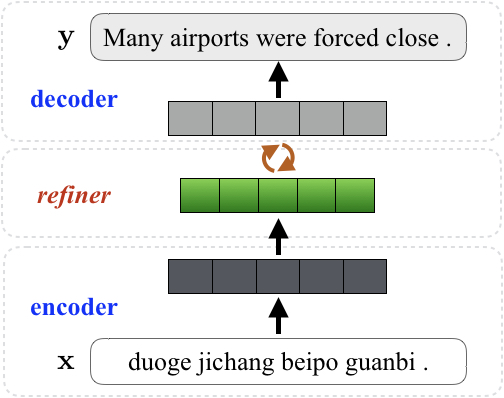}
\caption{Overview of the {\em encoder-refiner-decoder} NMT framework. The newly introduced \textit{refiner} dynamically updates the source representations based on the generated target-side information at each decoding step.}
\label{figure-overview}
\end{figure}

Inspired by human translation process, we propose a novel translation model, namely {\em encoder-refiner-decoder}, as illustrated in Figure~\ref{figure-overview}.
As seen, we introduce an additional {\em refiner} to dynamically refine the source representations by considering the target-side context information at each decoding time step. 
Specifically, the refiner consists of a {\em gate} that reads the target-side hidden state, the output of which is fed to a separate {\em re-encoder} to refine the source representations.
Since greedily refining source representations at every decoding step is time-consuming, 
we propose a conditional refining strategy, which adopts an auxiliary policy network trained by reinforcement learning to decide whether to perform the refine operation at each decoding step.

Experiments on Chinese--English and English--German corpora show that the proposed approach significantly improves translation performance by refining source representations for NMT. Furthermore, when refining strategy is applied, it alleviates the decoding speed problem by cutting down unnecessary refining operations. Results on Chinese--English translation task show improvement over NMT baseline systems of +2.34 BLEU points. As a trade-off, conditional refining strategy obtains a -0.51 BLEU point decrease but a substantive increase in decoding speed of approximately +33.33\%. 
Experiments for English--German translation task show a significant improvement of +1.23 BLEU points, demonstrating the potential universality of the proposed approach across language pairs. 

\paragraph{Contributions} Our main contributions can be summarized as follows:
\begin{enumerate}
    \item We proposed a novel {\em encoder-refiner-decoder} framework to produce target-aware source representations for improving NMT models;
    \item We introduce a policy network to reduce refining computations, making the approach highly practical, for example for translation in industry applications;
    \item We find our approach performs especially better on long sentences, which are generally complex and thus hard to be translated by NMT model. This finding confirm our claim that the introduced refiner can produce a better understanding of complex sentences.
\end{enumerate}

\section{Background}

Suppose that ${\bf x}=x_1, \dots, x_j, \ldots, x_J$ represents a source sentence and ${\bf y}=y_1, \dots, y_i, \dots, y_I$ a target sentence. NMT directly models the probability of translation from the source sentence to the target sentence word by word:
\begin{eqnarray}
P({\bf y}|{\bf x}) = \prod_{i=1}^{I} P(y_i| y_{<i}, {\bf x};\theta)
\end{eqnarray}
where $\theta$ is a set of model parameters and $y_{<i} = y_1, \dots, y_{i-1}$ denotes a partial translation before the position $i$.

The encoder-decoder architecture is now widely employed, where the encoder summarizes the source sentence ${\bf x}$ into a sequence of hidden states  ${\bf h} = h_1, \dots, h_j, \dots, h_J$ where $h_j$ is the hidden state of the $j$-th source word $x_j$, as in:
\begin{eqnarray}
{\bf h} = {encoder} (x_1, \dots, x_j, \dots, x_J)
\end{eqnarray}
in which $encoder$ is an encoding function to generate the a sequence of hidden states given all the related inputs. It can be either a Recurrent Neural Network (RNN)~\cite{hochreiter1997long,cho2014learning} or Convolutional Neural Network (CNN)~\cite{pmlr-v70-gehring17a} or Self-Attention Network (SAN)~\cite{Vaswani:2017:NIPS}. 
Next, the decoder generates each target word $y_i$ based on source context $c_i$, target context $s'$ and previously generated word(s) $y'$, as follows:
\begin{equation}\label{eq-refiner}
{s}_i = decoder(y', s', c_i)
\end{equation}
where $decoder$ is a decoding function to dynamically generate the decoder state $s_i$. The target word $y_i$ is generated given all the related inputs with non-linear activation function. Similar to $encoder$, it can be either a RNN, CNN or SAN. Besides, $s' \equiv s_{i-1}$ and $y' \equiv y_{i-1}$ in RNNsearch models, otherwise $s' \equiv s_{<i}$ and $y' \equiv y_{<i}$. $c_i$ is calculated by attention mechanism based on the source representations. 

Specifically, the source representations stand for source context, which embed the information (\eg syntax, semantics etc.) from the source sentence. However, in the traditional framework, the source representations always remain fixed during the whole decoding time steps regardless of target context. This actually adopts a greedy way to summarize excessive information (including relevant and irrelevant information) for generating each target word. Consequently, NMT needs to spend a substantial amount of its capacity in disambiguating source and target words based on the source context \cite{choi2017context}.

Therefore, it is necessary to take the target context into consideration to dynamically generate the target-aware source representations. Ideally, we expect the refined representations contain the information relevant to current target word, filtering irrelevant one.

\section{Approach}

\subsection{Refiner}

\begin{figure}[t]
\centering
\includegraphics[width=0.48\textwidth]{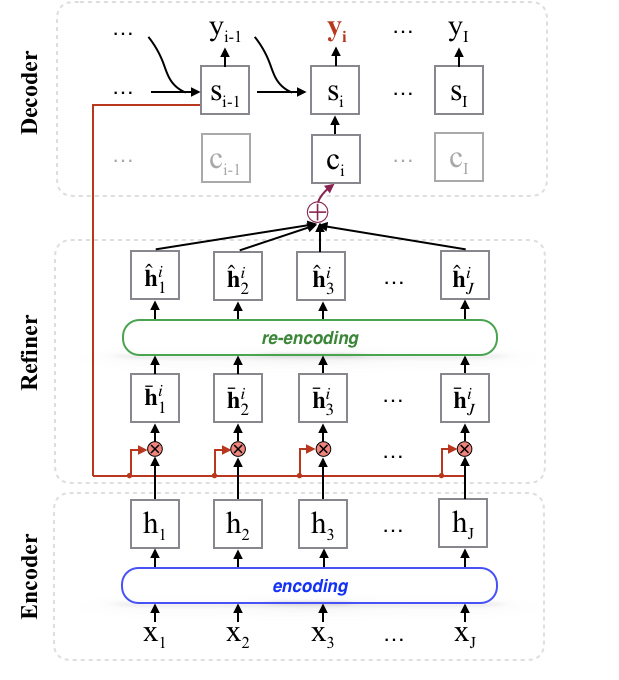}
\caption{Architecture of our {\em encoder-refiner-decoder} model. The original  source representations ${\bf h}$ produced by the standard encoder. The refined representations $\bar{\bf h}^i$ are generated by the context gate, which is conditioned on the target context $s_{i-1}$. Afterwards, we adopt an additional encoder to generate the re-understood representations $\hat{\bf h}^i$.}
\label{figure-architecture}
\end{figure}

As shown in Figure~\ref{figure-architecture}, the presented {\em encoder-refiner-decoder} framework literally consists of three major components: the standard {\em encoder}, {\em decoder} as well as the additional {\em refiner}. 
The basic idea of our approach is to dynamically refine source representation by using target context, and then use the decoder-sensitive representations as new source context for generating each target word. With the auxiliary context, we aim to encourage the source-side latent representations to embed dynamic target-side information, and thus generate better translation with enhanced representations.

\paragraph{Shallow Re-Understanding}

Originally, the attention mechanism is used to selectively summarize certain parts of source-side information. However, the source representations are fixed and thus treated universally for each target word. 
Accordingly, there may exist duplicated/useless information, which is irrelevant to the target word at current decoding step. 
From this observation, we introduce a context gate~\cite{Tu:2017:TACL} to dynamically control the amount of source representations used for generating the next target word at each time step.

At time step $i$, the refiner reads the previous hidden state of decoder (\eg ${s}_{i-1}$) and a sequence of hidden states from standard encoder (\ie ${\bf h}$), and then  refined hidden states (\ie $\bf{\bar h}^i$), which store decoder-sensitive information such as translated and untranslated contents~\cite{zheng2018modeling}. We introduce the context gate that consists of a sigmoid neural network layer and an element-wise multiplication operation. Each gate unit takes ${h}_j$ and ${s}_{i-1}$ to compute the gate vector $z^{i}_{j}$, and assigns an element-wise weight to ${h}_j$, computed by
\begin{align}
z^i_j & = \sigma (W_z h_{j} + U_z s_{i-1} + b_z) \label{eqn-context-gate} \\
{\bar h}^i_j & = z^{i}_j \odot h_j
\end{align}
where $\sigma(\cdot)$ is a sigmoid activation function and $\odot$ is element-wise multiplication. $W_z$ and $U_z$ are the weight matrices, and $b_z$ is the bias vector. These parameters are trained to learn how to refine source representation to maximize the overall translation performance. As results, we obtain the gated source representations ${\bf \bar h}^i$, which are associated with the decoder time step $i$.

\paragraph{Deep Re-understanding}

After adopt the target-side context into source representations via context gate, we then deeply re-understand them by an additional encoder. We expect this process can enhance the ability of distinguishing different translation predictions.
Given the tailored representations ${\bf \bar h}^i = {\bar h}^i_1, \dots, {\bar h}^i_j, \dots, {\bar h}^i_J$, we re-encode them to generate the refined representations $\hat{\bf h}^i_j$, as follows:
\begin{equation}
    {\bf \hat{h}^i} = {encoder}_{re}(\bar{h}^i_1, \dots, \bar{h}^i_j,\dots, \bar{h}^i_J)
\end{equation}
where ${encoder}_{re}$ is an encoding function, which is similar to ${encoder}$ in Equation (3) with different parameters. Furthermore, ${\bf \hat{h}^i}$ are used as a better source context for the decoder to generate the target word $y_i$ using the attention model. In experiment section, we verify the idea.

\subsection{Conditional Refining Strategy}

\paragraph{Definition}

The direct strategy is to process greedy refining at each decoder time step (\ie $1, \dots, i, \dots, I$). However, it is very time-consuming due to additional gating and re-encoding operations. One observation is that humans dynamically refresh the understanding of source sentence only after translating a complete semantic unit (\eg notional word, phrase, clause \etal). It is unnecessary to refine source representations at each decoder time step and some refining operations may be redundancy. Therefore, instead of greedy refining, we propose a conditional refining strategy to learn when to process refining.

Formally, at the decoder time step $i$, we parameterize the possible actions $a_i \in \{\textsc{refine},\textsc{reuse}\}$ with an auxiliary policy network, where \textsc{refine} indicates processing gating and re-encoding to refine representations (\ie ${\bf \hat{h}^{i+1}}$) while \textsc{reuse} represents skipping refining by reusing the refined source representations at previous time step (\ie ${\bf \hat{h}^{i}}$). We employ two-layer feed forward network to calculate the policy as follows:
\begin{equation}
    \pi(a_i|m_i) = softmax(W_p m_i + b_p)
\end{equation}
where $W_p$ is a weight matrix and $b_p$ is a bias vector. $m_i$ is a representation of current policy state, which is computed as:
\begin{equation}
    m_i = \tanh{(W'_p [s_i; E_{y_{i-1}}; c_i] + b'_p)}
\end{equation}
in which $s_i$ is the current decoder state, $E_{y_{i-1}}$ is the embedding of previous target word, and $c_i$ is the context vector. Note that our policy network makes the decision for the next decoding step rather than the current. 

\paragraph{Reinforcement Learning} 

Towards utilizing the discrete variables into the network, we employ the Gumbel-Softmax~\cite{jang2016categorical,maddison2016concrete} to approximate the one-hot vectors sampled from the categorical distribution with continuous representations. 
Using the reparameterization trick, the standard backpropagation can be utilized to compute the policy gradients of model parameters for reinforcement learning. 
As a result, the sample $\bar{\bf a}_i = (\bar{a}^1_i, \dots, \bar{a}^K_i) \in \mathbb{R}^K$ can be approximated using Gumbel-Softmax as follows:
\begin{align}
    \bar{\bf a}^k_i &= \frac{\exp((o^k_i + g^k_i)/\tau)}{\sum^{K}_{k^{\prime}=1} \exp((o^{k^{\prime}}_i + g^{k^{\prime}}_i)/\tau)} \\
    g^k_i &= -\log(-\log(u^k_i))\\
    u^k_i &\sim \text{Uniform}(0,1)
\end{align}
where $o_i$ is the unnormalized probability of $softmax$,  $g^k_i$ is {\em Gumbel noise}~\cite{gumbel1954statistical}, and $\tau \in (0, \infty)$ is a temperature parameter. 
The softmax function approaches argmax function as $\tau \rightarrow 0$, whereas it becomes uniform as $\tau \rightarrow \infty$.

In order to discretize the continuous probability $\bar{\bf a}_i$, we apply the straight-through version of Gumbel-Softmax, named Straight-Through (ST) Gumbel-Softmax~\cite{jang2016categorical}. 
During the forward phase, we use the Gumbel-Max trick, while computing the gradient with the continuous $\bar{\bf a}_i$.
Given the continuous probability $\bar{a}_i$ sampled from the Gumbel-Softmax, the discrete one-hot vector $\hat{a}_i = (\hat{a}^1_i, \dots, \hat{a}^K_i)$ are calculated as follows:
\begin{equation}
    \hat{a}^k_i = 
    \begin{cases}
    1 \quad k = \argmax_{k^{\prime}} \bar{a}^{k^{\prime}}_i\\
    0 \quad \text{otherwise}
    \end{cases}
\end{equation}

Using the above Gumbel-Softmax trick, at the decoder time step $i+1$, the choice on using the previous context-aware representations $\hat{\bf h}^i$ or the refined output $\hat{\bf h}^{i+1}$ can be formalized as follows:
\begin{equation}
    \tilde{\bf h}^{i+1} = \hat{a}^{\textsc{Reuse}}_{i} \hat{\bf h}^{i} + \hat{a}^{\textsc{Refine}}_{i} \hat{\bf h}^{i+1}
\end{equation}
where $\tilde{\bf h}^{i+1}$ is the final encoding representation for attention mechanism,  $\hat{a}^{\textsc{Reuse}}$ and $\hat{a}^{\textsc{Refine}}$ is the corresponding elements for the choices \textsc{Reuse} and \textsc{Refine} in the discrete one-hot vector $\hat{a}_{i}$.

Furthermore, we impose a constraint on the ratio of \textsc{Refine} operations to the total decoding length, encouraging the model to re-use the previous context-specific representations as much as possible.
As a remedy, we add a small penalty $r(\hat{a})$ the model pay for choosing the \textsc{Refine} operation as follows:
\begin{equation}
    r(\hat{a}) = \alpha (\sum^{I}_{i=1} \hat{a}^{\textsc{Refine}}_i) / I
\end{equation}
where $\alpha$ is hyper-parameter which controls the strength of penalty.

\section{Experiments}

\subsection{Data}

We carried out experiments on Chinese--English translation task. We used the corpus consisting of 1.25M bilingual sentence pairs extracted from LDC corpora.\footnote{The training corpus includes LDC2002E18, LDC2003E07, LDC2003E14, part of LDC2004T07, LDC2004T08 and LDC2005T06.} We used the NIST 2002 (MT02) as the tuning set for hyper-parameter optimization and model selection, and NIST 2003 (MT03), 2004 (MT04), 2005 (MT05), 2006 (MT06) and 2008 (MT08) as test sets. 
As seen, it totally contains 27.9M Chinese words and 34.5M English words. As most sentences in the corpus is in newswire domain, the average length is relatively longer than other informal domains (\eg 5.63 and 7.71 in subtitle corpus), which makes translation difficult.

We used case-insensitive 4-gram NIST BLEU metrics \cite{papineni2002bleu} as calcuated by the \verb|multi-bleu.perl|\footnote{\url{https://github.com/moses-smt/mosesdecoder}} script for evaluation, and {\em sign-test} \cite{Collins05} to test for statistical significance.

\begin{table*}[ht]
  \centering
  \small
  \renewcommand{\arraystretch}{1.3}
  \begin{tabular}{l|c|cccccc|c}
    \bf Model & \bf MT02  &  \bf  MT03 & \bf MT04 & \bf MT05 & \bf MT06 & \bf MT08 & \bf Ave. & \bf $\bigtriangleup$\\
    \hline
    Baseline & 40.16 &  37.26  &  40.50  &  36.67  &  37.10  & 28.54 & 36.01 & --\\
    ~~ + Multi-layer & 40.18 & 37.30 & 40.75 & 37.43 & 37.29 & 28.72 & 36.30 & +0.29\\
    \hline
    ~~ + Shallow Refiner &   41.08   &   38.12$^{\dag}$    &   41.35$^{\dag}$    &   38.51$^{\dag}$    &   37.85$^{\dag}$   &  29.32$^{\dag}$ & 37.03 & +1.02\\
    ~~ + Deep Refiner &  41.58  & 39.23$^{\dag}$   &  42.72$^{\dag}$   &  39.90$^{\dag}$   &  39.24$^{\dag}$   & 30.68$^{\dag}$ &  38.35 & \bf +2.34\\
    \hdashline
    ~~ + Conditional & 41.02 & 38.73$^{\dag}$ & 42.34$^{\dag}$ & 39.23$^{\dag}$ & 38.92$^{\dag}$ & 29.97$^{\dag}$ & 37.84 & +1.83\\
  \end{tabular}
  \caption{Evaluation of translation performance for Chinese--English. ``Baseline'' is trained with standard NMT model, while ``+ Multi-layer'' is trained with an additional encoder layer. ``+ Shallow Refiner'' indicates adding context gate for refining source representations, and ``+ Deep Refiner'' denotes processing re-encoding on refined representations. ``+ Contiditonal'' indicates the conditional refining strategy to decide when to refine the source representations. ``$\bigtriangleup$'' column denotes performance improvements over ``Baseline''. ``$\dag$''indicate statistically significant difference ($p < 0.01$) from baseline.}
  \label{tab-results}
\end{table*}

\subsection{Model}

The baseline is our re-implemented attention-based NMT system, which incorporates dropout \cite{hinton2012improving} on the output layer and improves the attention model by feeding the most recently generated word. For training the translation models, we limited the source and target vocabularies to the most frequent 30K words in Chinese and English, covering approximately 97.2\% and 99.3\% of the words in the two languages, respectively. Our models were trained on sentences of length up to a maximum of 50 words with early stopping. Mini-batches were shuffled during processing with a mini-batch size of 80. The word-embedding dimension was 620 and the hidden layer size was 1,000. We set learning rate as $5 \times {10}^{-4}$, gradient norm as 1.0 and dropout rate as 0.3. We applied {\em Rmsprop}~\cite{graves2013generating} to train models for 10 epochs and selected the model parameters that yielded best performances on the tuning set. The beam size is set as 10. The proposed model was implemented on top of the baseline model with the same settings where applicable. The hidden layer size in the refiner was 1,000. 

\subsection{Results and Discussion}

We evaluated the presented approaches in terms of translation quality, speed and robustness.

\begin{table}[t]
  \centering
  \small
  \renewcommand{\arraystretch}{1.3}
  \begin{tabular}{l|r|r|c}
    \bf Model & \bf \# Params & \bf Speed & \bf P \% \\
    \hline
    Baseline & 86.69M &  558  & -- \\ 
    ~~ + Multi-layer & 104.70M & 516 & --\\
    \hline
    ~~ + Shallow Refiner & 92.69M & 499 & 100 \\
    ~~ + Deep Refiner & 110.70M & 132 & 100 \\
    \hdashline
    ~~ + Conditional & 112.95M & 176 & 59
  \end{tabular}
  \caption{Decoding speed on Chinese--English translation task. ``Params'' denotes the number of parameters.  ``Speed'' is measured in words/second. ``P'' indicates the percentage of refining operations over the translated words.}
  \label{tab-speed}
\end{table}

\paragraph{Translation Quality}
Table~\ref{tab-results} shows translation performances for Chinese--English.
The two baseline NMT models, one being trained with the standard NMT system (\ie ``{Baseline}''), while the other was trained with multi-layer encoders (\ie ``+ Multi-layer''). Benefiting from the deeper layers, the stronger baseline system is able to improve performance over the standard baseline system by + 0.29 BLEU point.

Clearly the proposed models significantly improve the translation quality in all cases, although there are still considerable differences among different variants.
Introducing context gate for refining source representations (\ie ``Shallow Refiner'') improves translation performance over ``Baseline'' by +1.02 BLEU points. 
It demonstrates the effectiveness of our proposed refiner model over the baseline model.
Furthermore, adding re-encoding (\ie ``Deep Refiner'') together achieves the best performance overall, which is +2.34 BLEU points better than the baseline model.
This confirms our assumption that re-encoder applied to the shallow-refined source representations indeed help to re-understand the deeper semantics of source sentence. 
The ``Conditional Strategy'' shows the the policy network can skip 41\% unnecessary refining operations (as illustrated in Table~\ref{tab-speed}) but still keep reasonable translation performances (\ie around +1.83 BLEU points than ``Baseline''). {{Note that we can control the percent of refining operations depending on the requirement via the hyper-parameter $\alpha$ and threshold value of choosing the refining operation. In this work, we choose appropriate $\alpha$ and threshold value to report the corresponding result.}}

\paragraph{Parameters}

In terms of additional parameters introduced by the refining models, both shallow and deep refiners introduce a large number of parameters. Beginning with the baseline model's 86.69M parameters, the ``+ Shallow Refiner'' adds +6.00M new parameters, while the ``+ Deep Refiner'' adds a further +18.01M new parameters with an additional encoder layer. For fair comparison, our ``+ Multi-layer'' also adds +18.01M new parameters by adding same encoder layer over ``Baseline'' as a stronger baseline. Besides, ``+ Conditional'' needs further +2.25M new parameters to learn the decision of skipping refining. 

\paragraph{Speed}
More parameters may capture more information, at the cost of posing difficulties to training. Although gains are made in terms of translation quality by introducing refining, we need to consider the potential trade-off with respect to a possible increase in time consumption, due to the large number of newly introduced parameters resulting from the incorporation of context gate and additional encoder into the NMT model. As shown in Table~\ref{tab-speed}, when running on a single GPU device Tesla P40, the decoding speed of ``Baseline'' is 558 target words per second, and this reduces to 499 words per second with a slight decrease when context gate is added. With the introduction of additional encoder, the decoding speed has a drastic decrease to 132 target words per second. In terms of decoding time trade-off, our conditional refining model increases decoding speed by 33.33\% over the ``+ Deep Refiner'' . We attribute this to the fact that no re-encoding for each step, which avoids high computation consumption. {{Taking the time consumption into consideration, our ``+ Shallow Refiner'' can be utilized in online decoding scenario while ``+ Deep Refiner'' is more appropriate for offline translation. Further, we can leverage the conditional strategy to balance the decoding speed and translation performance.}}

\begin{table}[t]
\centering
\small
\renewcommand\arraystretch{1.3}
\begin{tabular}{l|c|c|c }
	\bf Model  & \bf news-2013 & \bf news-2014 & \bf $\bigtriangleup$\\
	\hline
	{Baseline} & 22.35 & 22.33 & --\\
	~~ + Multi-layer & 22.32 & 22.39 & +0.06\\
	\hline
	~~ + Shallow Refiner & 22.95 & 23.02$^{\dag}$ & +0.69\\
	~~ + Deep Refiner  & 23.20 & 23.56$^{\dag}$ & \bf +1.23\\
	\hdashline
	~~ + Conditional & 23.06 & 23.27$^{\dag}$ & +0.94\\
\end{tabular}
\caption{\label{tab-results-ende} Translation performance on English--German.}
\end{table}

\paragraph{English--German Translation Task}

To validate the robustness of our approach on other language pairs, we conducted experiments on WMT2014 English-German corpus, which contains 4.5M bilingual sentence pairs with 116M English words and 110M German words.\footnote{\url{http://www.statmt.org/wmt14/}} We use {\em news-test2013} as tuning set and the {\em news-test2014} as test set. Particularly, we segmented words via byte pair encoding (BPE)~\cite{sennrich2016neural}. We consider a joint source and target byte-pair encoding with 32K types. We set beam size as 4 in our work. We used the same settings as used in Chinese--English experiments. 
As shown in Table \ref{tab-results-ende}, our proposed ``+ Shallow Refiner'' and ``+ Deep Refiner'' also significantly improves translation performance on the English--German task, demonstrating the efficiency and universality of the proposed approach.

\subsection{Analysis}

\begin{figure}[t]
\centering
\includegraphics[width=0.35\textwidth]{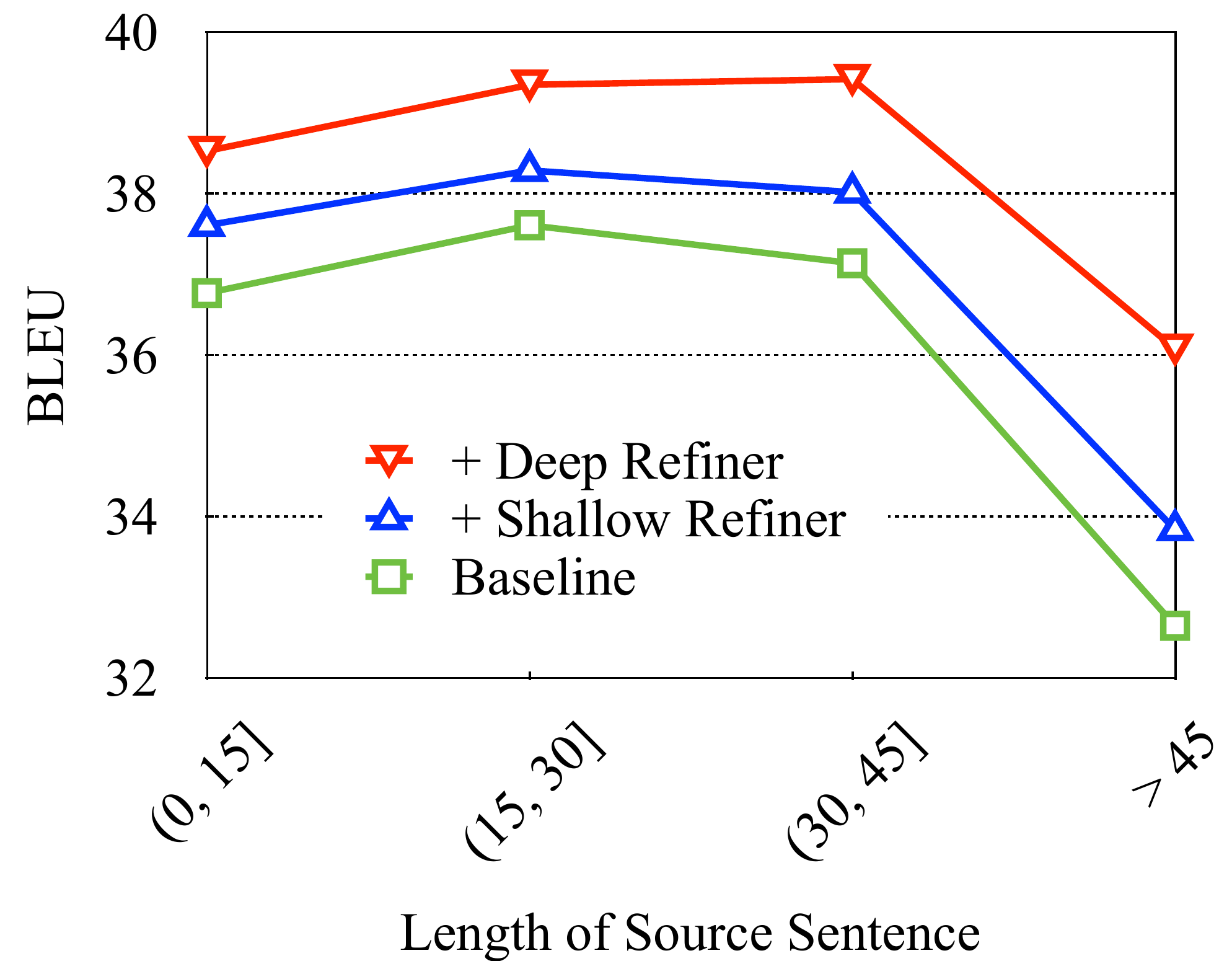}
\caption{Performance of the generated translations with respect to the lengths of the source sentences.}
\label{figure-length}
\end{figure}

\paragraph{Length Analysis}
We follow \citeauthor{tu2016modeling}~\shortcite{tu2016modeling} to group sentences with similar lengths together.
As shown in Figure~\ref{figure-length}, our ``+ Shallow Refiner'' and ``+ Deep Refiner'' substantially outperform the ``Baseline'' on each length span.
Moreover, ``+ Deep Refiner'' also makes the remarkable improvement over ``+ Shallow Refiner'' on the entire length segments. 
More importantly, we discover that the increment percent of ``+ Shallow Refiner'' and ``+ Deep Refiner'' grows drastically, as the length of source sentences rises. 
Specifically, the ``+ Shallow Refiner'' on the length span (\eg $\geq$ 45) increases 3.64\% BLEU over the ``Baseline''.
As a comparison, the ``+ Deep Refiner'' start with 4.79\% increment(\eg $<$ 15), and keep the upward trend, finally obtaining excellent 10.50\% growth (\eg $\geq$ 45).
The significant improvements of our refiner-based models can be attributed to dynamically re-understanding the source sentence based on the target-side context.
Especially, when to deal with the complex sentences, our refiner-based models can better capture the context information related to target context than the standard models.
Furthermore, we observe that the increment percent ``+ Deep Refiner'' is more obvious that ``+ Shallow Refiner'' on long source sentences. 
From this fact, we argue that the necessity of deep re-understanding with additional encoder for the complex sentences.

\paragraph{Effects on Linguistic Insights}

Towards investigating the distributions of the learned refining policy, we evaluate the consistency between linguistic categories\footnote{We utilized OpenNLP toolkit (\url{https://opennlp.apache.org/}) to automatically annotated the translation output with chunk tags.} (\ie chunk) and the {\textsc{refine}} operation. We measured the consistency by calculating the percentage of {\textsc{refine}} operations in one corresponding type, and the results are shown in Table~\ref{tab-chunk-pos}.

As seen, the {\textsc{refine}} operation at the beginning position (\ie ``B'') of chunks happens more times than that at other positions (\ie ``I''). Taking NP for example, 67.67\% refining happens at the beginning of chunks while 52.60\% at the following positions. This confirms our assumption that there is no need to refine the units inside a semantic component such as phrase and clause. Beside, we found a contrary phenomenon in VPs, where ``I\%'' is bigger than ``B\%''. The reason may be that fictional words that have little lexical meaning often occurs at begging positions such as ``has {\em made}'' and ``to {\em severely punish}'', and it is unnecessary to process refining on these kinds of words.

\begin{table}[t]
\centering
\small
\renewcommand\arraystretch{1.1}
\begin{tabular}{r|r|r r}
	\bf Type & \bf Overall \% & \bf B \% & \bf I \%\\
	\hline
	NP & 42.28 &  67.67   &   52.60 \\
	PP & 18.56 &  63.18   &   58.82 \\
	VP & 15.24 &  62.91   &   72.62 \\
	ADVP & 2.13 & 72.21   &   62.96 \\
	ADJP & 0.90 & 70.75   &   61.29 \\
	\hdashline
	Others & 20.89 & 80.80 & 66.67\\
\end{tabular}
\caption{\label{tab-chunk-pos} Relations between refining policy and chunking categories. ``Overall'' donates the percentage of the type of words over the whole data set. ``B'' and ``I'' donate the beginning and other parts in corresponding chunk, respectively.}
\end{table}

\paragraph{Effect of Gated Refiner}

Some researchers may argue the gated refiner computes a scalar as output, rather than the vector as ours. 
We force the gate refiner to compute the scalar weight as output (\ie ``Hard Mask'').
As a result, the corresponding ``+ Shallow Hard Mask'' and ``+ Deep Hard Mask '' achieves comparable performances with a slight improvement about +0.11 and +0.21 BLEU over ``{Baseline}'', but significant decreases about -0.91 and -2.13 BLEU over our proposed ``+ Shallow Refiner'' and ``+ Deep Refiner''. 
We conjecture that the above results attribute to the more expressive capability of the vector that scalar output.
In our work, the target of context gate assigns the weight to source representations.
We assume that the each element in the source representation is taken as a feature, and the output of context gate is the corresponding weight, indicating the importance.
With the vector as the weight, we assign large scores for some important features but small for others.
Therefore, from this viewpoint, it's reasonable that our context gate achieves significant improvement over the scalar counterparts.

\paragraph{Case Study}

\begin{figure}[t]
\centering
\includegraphics[width=0.4\textwidth]{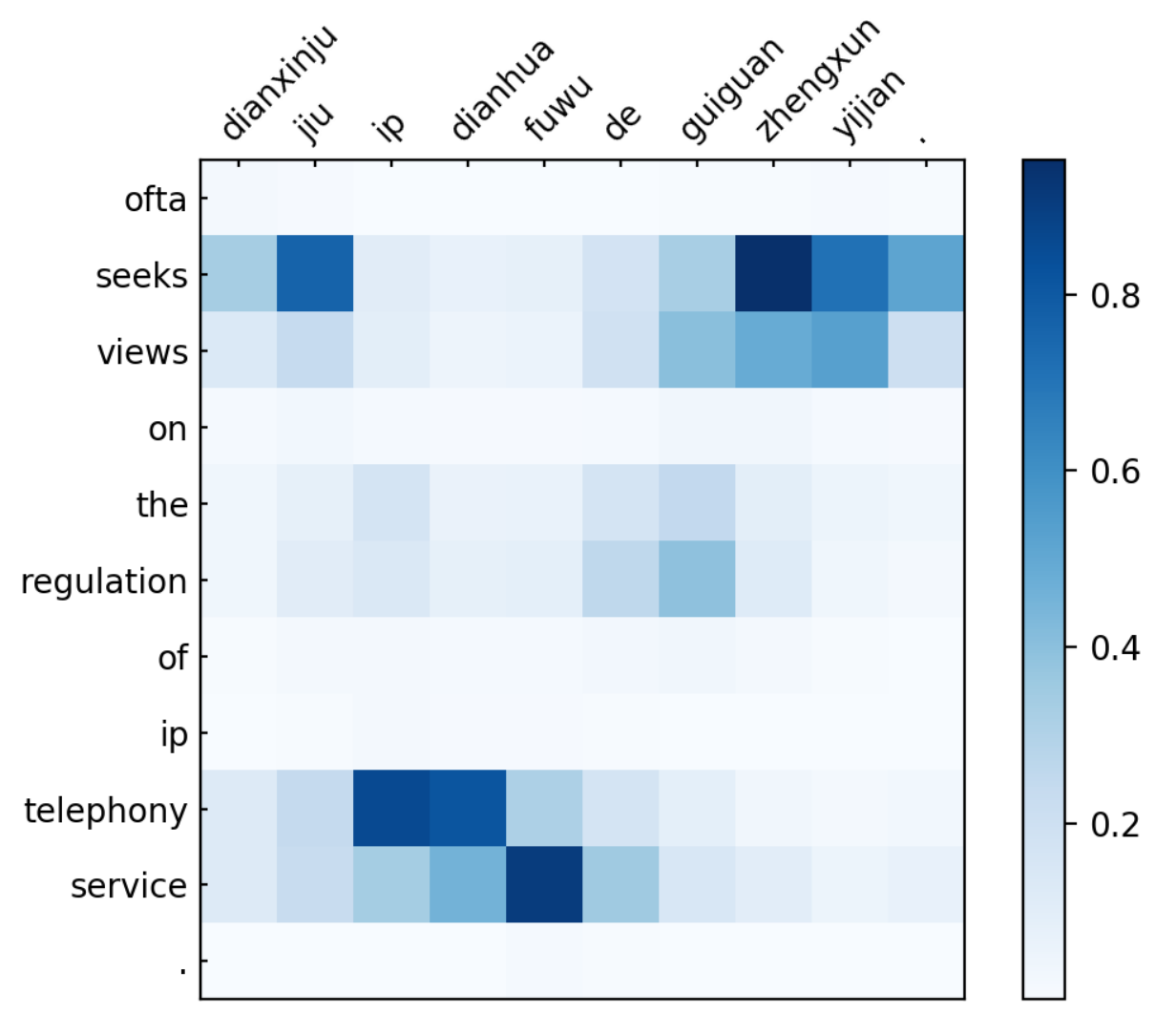}
\caption{First derivative heat map of the output with respect to the context gate vector.}
\label{figure-vis}
\end{figure}

Towards investigating the refining over source tokens based target partial translation, we visualize the corresponding context gate for each decoding time step.
Motivated by visualizing neural model strategy in \citeauthor{li2015visualizing}\shortcite{li2015visualizing}, the contributions of gating vector to final output can be approximated by first derivatives.
At each decoding time step, we compute first derivatives for gating vector with back-propagation to measure saliency score. 
As illustrated in Figure~\ref{figure-vis}, our model can select relevant important fragments in the source sentence, taking the current decoding context into consideration.
For example, when generating the target words ``seeks'', the relevant fragments ``zhengxun'' and ``yijian'' in source sentence attach great importance.
Although the saliency scores of source tokens ``dianxinju'' and ``ip'' are extremely low during the generation of the target tokens ``ofta'' and ``ip'', the aforementioned scores are highest among all the source tokens.
In contrast, when the decoder predicts the target meaningless prepositions ``on'' and ``of'', the corresponding gating mechanism prefers to paying no attention over the source sentence. 
That is, when generating the target prepositions, the information of source sentence is useless for current decoding. 
Hence, the generation of target prepositions only relies on preceding target context.
These results verify the proposal that generation of a content word should rely more on the source context and generation of a functional word should rely more on the target context~\cite{Tu:2017:TACL}.

\begin{table}[t]
\centering
\small
\renewcommand\arraystretch{1.3}
\begin{tabular}{l|c|c}
	\bf Model & \bf BLEU & \bf $\bigtriangleup$ \\
	\hline
	Baseline            &   36.01   &   --\\
	\hline
	~~+ Shallow Hard Mask &   36.12   &   +0.11\\
	~~+ Deep Hard Mask &   36.22   &   +0.21\\
	\hline
	~~+ Shallow Refiner &   37.03   &   +1.02\\
	~~+ Deep Refiner &   38.35   &   +2.34\\
\end{tabular}
\caption{\label{tab-results-no-rec} Evaluation of translation performance when {\em Hard Mask} is used as context gate. ``BLEU'' is calculated as the average BLEU score on test sets.}
\end{table}

\section{Related Work}

\paragraph{Conditional Sequence Processing}
More relevant to our work, \citeauthor{Ke:2018:ICML}~\shortcite{Ke:2018:ICML} present a focused hierarchical RNNs architecture for sequence modelling tasks, which allows them to attend to key parts of the
input as needed.
Similarly, a discrete gating mechanism is adopted to make a discrete decision on whether or not the token is relevant to current context. Subsequently, the selected tokens are feed into the high RNN layer.
Essentially the gating mechanism is hard mask, trained using reinforcement learning.
In comparison with their work, our context gate computes a gate vector to mask the corresponding source vector. 
More importantly, it's proven that the introduction of the hard mask into our work has no effect, showing a substantial margin with our context gate in our experimental results.
In addition, our model can be optimized using the standard methods instead of reinforcement learning. 

\paragraph{GRU-Gated Attention}
Recently, \citeauthor{zhang2017gru}~\shortcite{zhang2017gru} proposed a gru-gated attention to consider the decoding context into calculating the context vector with attention mechanism. 
Similar to our adopted context gate module, \citeauthor{zhang2017gru}~\shortcite{zhang2017gru} also introduces a gating layer to refine the source representations at each decoding time step, based on the current decoding state. 
Instead of using conventional gating mechanism, a GRU cell is chose to deal with complex interactions between source sentence and partial translation. 
Furthermore, another important difference from their work is that our work subsequently adopts the additional encoder to further encode the tailored decoding context-aware source representations again, which has been proven its excellent effectiveness in corresponding experimental results.

\paragraph{Context-Dependent Word Embedding} More than one meaning of a word can be encoded through measuring the multiple dimensions of similarity. 
Towards explicitly disambiguating source and target words, \citeauthor{choi2017context}~\shortcite{choi2017context} propose to contextualize the word embedding vectors using a nonlinear bag-of-words representation of the source sentence.
Similarly, we propose to learn to refine the continuous representations to generate target-side context-aware representations. 
However, different with the source context-dependent word embedding, our proposed {\em encoder-refiner-decoder} refines the source sentence representations generated by standard encoder rather than word embedding to produce the target-side context-aware representations for each decoding time.

\paragraph{Novel Model Architectures}
~\citeauthor{Tu:2017:AAAI}~\shortcite{Tu:2017:AAAI} introduced a novel {\em encoder-decoder-reconstructor} architecture, which reconstructs decoder states back to the original input sentence. ~\citeauthor{Wang:2018:AAAI}~\shortcite{Wang:2018:AAAI,Wang:2018:EMNLP} moved one step further by simultaneously reconstructing encoder states back to the input sentence.
~\citeauthor{Xia:2017:NIPS}~\shortcite{Xia:2017:NIPS} and~\citeauthor{Zhang:2018:AAAI}~\shortcite{Zhang:2018:AAAI} independently introduced a {\em second-pass decoder} to polish the raw translation generated by the first-pass decoder. In contrast, we introduced a refiner to polish the encoder states, and thus can be regarded as a {\em second-pass encoder}.

\section{Conclusion and Future Work}

This paper is an early attempt to use target context to refine source representations for improving translation performance. 
As a result, the generated source representations concentrate on most relevant semantic to current target-side context. 
Furthermore, as a trade-off between efficiency and effectiveness, we further propose to learn when to refine the source representations at each decoding time step.
We train the policy network for learning to refine with reinforcement learning.
The experimental results on Chinese--English and English--German translation tasks demonstrate the remarkable effectiveness over the standard encoder-decoder architecture.
In addition, it's proven that the excellent performance of our proposed architecture on the long sentences. In future work we plan to adopt a diversity of context information except for target context into our proposed architecture to improve the translation.

\balance
\bibliography{ijcai19}
\bibliographystyle{named}

\end{document}